\documentclass{article}

\usepackage{etoolbox}
\newtoggle{arxiv}
\toggletrue{arxiv}

\usepackage[utf8]{inputenc} 
\usepackage[T1]{fontenc}    
\usepackage{hyperref}       
\usepackage{url}            
\usepackage{booktabs}       
\usepackage{amsfonts}       
\usepackage{nicefrac}       
\usepackage{microtype}      
\usepackage{xcolor}         
\usepackage{amsmath,amsthm,amssymb}

\usepackage{algorithm}
\usepackage{algorithmic}

\usepackage{subcaption}
\usepackage{multirow}

\usepackage{xspace}
\usepackage{wrapfig}

\usepackage{pifont}

\usepackage[capitalise]{cleveref}  
\usepackage{comment}
\usepackage[inline]{enumitem}

\usepackage{graphicx}
\usepackage{grffile}
\usepackage{color}

\usepackage{newtxmath}

\iftoggle{arxiv}{
\usepackage[numbers,sort]{natbib}
}{}

\usepackage{import}

\newtoggle{todo}
\toggletrue{todo}

\newtoggle{comment}
\togglefalse{comment} 

\newcommand{\diag}{\mathrm{diag}}
\newcommand{\softmax}{\mathrm{softmax}}
\newcommand{\dsoftmax}{\mathrm{dsoftmax}}

\newcommand{\vQ}{\mathbf{Q}}
\newcommand{\vK}{\mathbf{K}}
\newcommand{\vV}{\mathbf{V}}
\newcommand{\vdQ}{\mathbf{dQ}}
\newcommand{\vdK}{\mathbf{dK}}
\newcommand{\vdV}{\mathbf{dV}}
\newcommand{\vS}{\mathbf{S}}
\newcommand{\vdS}{\mathbf{dS}}
\newcommand{\vP}{\mathbf{P}}
\newcommand{\vdP}{\mathbf{dP}}

\newcommand{\vO}{\mathbf{O}}
\newcommand{\vdO}{\mathbf{dO}}

\newcommand{\sysnameone}{\textsc{FlashAttention}\xspace}
\newcommand{\sysname}{\textsc{FlashAttention-2}\xspace}

\newtheorem*{theorem*}{Theorem}

\iftoggle{arxiv}{
  \setlength{\textwidth}{6.5in}
  \setlength{\textheight}{9in}
  \setlength{\oddsidemargin}{0in}
  \setlength{\evensidemargin}{0in}
  \setlength{\topmargin}{-0.5in}
  \newlength{\defbaselineskip}
  \setlength{\defbaselineskip}{\baselineskip}
  \setlength{\marginparwidth}{0.8in}
}{
\usepackage[compact]{titlesec}
\titlespacing{\section}{0pt}{*1}{*0}
\titlespacing{\subsection}{0pt}{*1.5}{*0}

\usepackage[subtle, mathdisplays=tight, charwidths=tight, leading=normal]{savetrees}

\addtolength\textfloatsep{-0.5em}
\addtolength\intextsep{-0.2em}

\def\setstretch#1{\renewcommand{\baselinestretch}{#1}}
\setstretch{0.985}
\addtolength{\parskip}{-1pt}

}

\title{\sysname:\\Faster Attention with Better Parallelism and Work Partitioning}

\iftoggle{arxiv}{
  \usepackage{authblk}
  \author[1,2]{Tri Dao}
  \affil[1]{Department of Computer Science, Princeton University}
  \affil[2]{Department of Computer Science, Stanford University}
  \affil[ ]{\texttt{trid@cs.stanford.edu}}
}{
}

\begin{document}

\maketitle

\begin{abstract}
Scaling Transformers to longer sequence lengths has been a major problem in the
last several years, promising to improve performance in language modeling and
high-resolution image understanding, as well as to unlock new applications in
code, audio, and video generation.
The attention layer is the main bottleneck in scaling to longer sequences, as
its runtime and memory increase quadratically in the sequence length.
\sysnameone~\citep{dao2022flashattention} exploits the asymmetric GPU memory
hierarchy to bring significant memory saving (linear instead of quadratic) and
runtime speedup (2-4$\times$ compared to optimized baselines), with no approximation.
However, \sysnameone is still not nearly as fast as optimized matrix-multiply
(GEMM) operations, reaching only 25-40\% of the theoretical maximum FLOPs/s.
We observe that the inefficiency is due to suboptimal work partitioning between
different thread blocks and warps on the GPU, causing either low-occupancy or
unnecessary shared memory reads/writes.
We propose \sysname, with better work partitioning to address these issues.
In particular, we (1) tweak the algorithm to reduce the number of non-matmul
FLOPs (2) parallelize the attention computation, even for a single head, across
different thread blocks to increase occupancy, and (3) within each thread block,
distribute the work between warps to reduce communication through shared memory.
These yield around 2$\times$ speedup compared to \sysnameone, reaching 50-73\% of the
theoretical maximum FLOPs/s on A100 and getting close to the efficiency of GEMM
operations.
We empirically validate that when used end-to-end to train GPT-style models,
\sysname reaches training speed of up to 225 TFLOPs/s per A100 GPU (72\% model
FLOPs utilization).\footnote{\sysname
  is available at \url{https://github.com/Dao-AILab/flash-attention}}


\end{abstract}

\section{Introduction}
\label{sec:intro}

Scaling up the context length of Transformers~\citep{vaswani2017attention} is a
challenge, since the attention layer at their heart has runtime and
memory requirements quadratic in the input sequence length.
Ideally, we would like to go beyond the standard 2k sequence length limit to
train models to understand books, high resolution images, and long-form videos.
Just within the last year, there have been several language models with much
longer context than before: GPT-4~\citep{OpenAI2023GPT4TR} with context
length 32k, MosaicML's MPT with context length 65k, and Anthropic's
Claude with context length 100k.
Emerging use cases such as long document querying and story writing have
demonstrated a need for models with such long context.

To reduce the computational requirement of attention on such long context, there
have been numerous methods proposed to approximate
attention~\citep{kitaev2020reformer, roy2021efficient, wang2020linformer,
  katharopoulos2020transformers, choromanski2020rethinking,
  beltagy2020longformer, zaheer2020bigbird, scatterbrain}.
Though these methods have seen some use cases, as far as we know, most
large-scale training runs still use standard attention.
Motivated by this, \citet{dao2022flashattention} proposed to reorder the
attention computation and leverages classical techniques (tiling, recomputation)
to significantly speed it up and reduce memory usage from quadratic to linear in
sequence length.
This yields 2-4$\times$ wall-clock time speedup over optimized baselines, up to
10-20$\times$ memory saving, with no approximation, and as a result \sysnameone has
seen wide adoption in large-scale training and inference of Transformers.

However, context length increases even more, \sysnameone is still not nearly as
efficient as other primitives such as matrix-multiply (GEMM).
In particular, while \sysnameone is already 2-4$\times$ faster than a standard
attention implementation, the forward pass only reaches 30-50\% of the
theoretical maximum FLOPs/s of the device (\cref{fig:benchmark_attn_fwd}), while
the backward pass is even more challenging, reaching only 25-35\% of maximum
throughput on A100 GPU (\cref{fig:benchmark_attn_bwd}).
In contrast, optimized GEMM can reach up to 80-90\% of the theoretical maximum
device throughput.
Through careful profiling, we observe that \sysnameone still has suboptimal work
partitioning between different thread blocks and warps on the GPU, causing
either low-occupancy or unnecessary shared memory reads/writes.

Building on \sysnameone, we propose \sysname with better parallelism and work
partitioning to address these challenges.
\begin{enumerate}
  \item In \cref{subsec:algo}, we tweak the algorithms to reduce the number of non-matmul FLOPs while not
  changing the output.
  While the non-matmul FLOPs only account for a small fraction of the total FLOPs,
  they take longer to perform as GPUs have specialized units for matrix multiply,
  and as a result the matmul throughput can be up to 16$\times$ higher than non-matmul
  throughput.
  It is thus important to reduce non-matmul FLOPs and spend as much time as
  possible doing matmul FLOPs.

  \item We propose to parallelize both the forward pass and backward pass along
  the sequence length dimension, in addition to the batch and number of heads
  dimension. This increases occupancy (utilization of GPU resources) in the case
  where the sequences are long (and hence batch size is often small).

  \item Even within one block of attention computation, we partition the work
  between different warps of a thread block to reduce communication and shared
  memory reads/writes.

\end{enumerate}

In \cref{sec:experiments}, we empirically validate that \sysname yields significant speedup compared to
even \sysnameone. Benchmarks on different settings (with or without causal mask,
different head dimensions) show that \sysname achieves around 2$\times$ speedup over
\sysnameone, reaching up to 73\% of the theoretical max throughput in the
forward pass, and up to 63\% of the theoretical max throughput in the backward pass.
When used end-to-end to train GPT-style models, we reach training speed of up to
225 TFLOPs/s per A100 GPU.


\section{Background}
\label{sec:background}

We provide some background on the performance characteristics and execution
model of GPUs.
We also describe the standard implementation of attention, as well as
\sysnameone.

\subsection{Hardware characteristics}
\label{subsec:hardware}

\textbf{GPU performance characteristics.}
The GPU consists of compute elements (e.g., floating point arithmetic units) and
a memory hierarchy.
Most modern GPUs contain specialized units to accelerate matrix multiply in
low-precision (e.g., Tensor Cores on Nvidia GPUs for FP16/BF16 matrix multiply).
The memory hierarchy comprise of high bandwidth memory (HBM), and on-chip SRAM
(aka shared memory).
As an example, the A100 GPU has 40-80GB of high bandwidth memory (HBM) with
bandwidth 1.5-2.0TB/s and 192KB of on-chip SRAM per each of 108 streaming
multiprocessors with bandwidth estimated around 19TB/s~\citep{jia2018dissecting,
  jia2021dissecting}.
As the L2 cache is not directly controllable by the programmer, we focus on the
HBM and SRAM for the purpose of this discussion.

\textbf{Execution Model.}
GPUs have a massive number of threads to execute an operation
(called a kernel).
Threads are organized into thread blocks, which are scheduled to run on
streaming multiprocessors (SMs).
Within each thread blocks, threads are grouped into warps (a group of 32
threads).
Threads within a warp can communicate by fast shuffle instructions or cooperate
to perform matrix multiply.
Warps within a thread block can communicate by reading from / writing to shared memory.
Each kernel loads inputs from HBM to registers and SRAM, computes, then writes outputs to HBM.

\subsection{Standard Attention Implementation}
\label{subsec:standard_attn}

Given input sequences $\vQ, \vK, \vV \in \mathbb{R}^{N \times d}$ where $N$ is the sequence length and
$d$ is the head dimension, we want to compute the attention output $\vO \in \mathbb{R}^{N \times d}$:
\begin{equation*}
  \vS = \vQ \vK^\top \in \mathbb{R}^{N \times N}, \quad \vP = \softmax(\vS) \in \mathbb{R}^{N \times N}, \quad \vO = \vP\vV \in \mathbb{R}^{N \times d},
\end{equation*}
where $\softmax$ is applied row-wise.\footnote{For clarity of exposition, we
  omit the scaling of $\vQ \vK^\top$ (typically by $1/\mathrm{d}$), and optionally
  elementwise masking on $\vS$ and/or dropout applied to $\vP$} For multi-head
attention (MHA), this same computation is performed in parallel across many
heads, and parallel over the batch dimension (number of input sequences in a
batch).

The backward pass of attention proceeds as follows.
Let $\vdO \in \mathbb{R}^{N \times d}$ be the gradient of $\vO$ with respect to some loss
function. Then by the chain rule (aka backpropagation):
\begin{align*}
  \vdV &= \vP^\top \vdO \in \mathbb{R}^{N \times d} \\
  \vdP &= \vdO \vV^\top \in \mathbb{R}^{N \times N} \\
  \vdS &= \dsoftmax (\vdP) \in \mathbb{R}^{N \times N} \\
  \vdQ &= \vdS \vK \in \mathbb{R}^{N \times d} \\
  \vdK &= \vQ \vdS^\top \in \mathbb{R}^{N \times d},
\end{align*}
where $\dsoftmax$ is the gradient (backward pass) of softmax applied row-wise.
One can work out that if $p = \softmax(s)$ for some vector $s$ and $p$, then
with output gradient $dp$, the input gradient $ds = (\diag(p) - p p^\top)dp$.

Standard attention implementations materialize the matrices $\vS$ and $\vP$ to
HBM, which takes $O(N^2)$ memory.
Often $N \gg d$ (typically $N$ is on the order of 1k--8k and $d$ is around 64--128).
The standard attention implementation (1) calls the matrix multiply (GEMM)
subroutine to multiply $\vS = \vQ \vK^\top$, writes the result to HBM, then (2)
loads $\S$ from HBM to compute softmax and write the result $\vP$ to HBM, and
finally (3) calls GEMM to get $\vO = \vP \vV$.
As most of the operations are bounded by memory bandwidth, the large number of
memory accesses translates to slow wall-clock time.
Moreover, the required memory is $O(N^2)$ due to having to materialize $\vS$ and
$\vP$.
Moreover, one has to save $\vP \in \mathbb{R}^{N \times N}$ for the backward pass to compute the
gradients.

\subsection{\sysnameone}
\label{subsec:flashv1}

To speed up attention on hardware accelerators such as GPU,
\citep{dao2022flashattention} proposes an algorithm to reduce the memory
reads/writes while maintaining the same output (without approximation).

\subsubsection{Forward pass}
\sysnameone applies the classical technique of tiling to reduce memory IOs, by
(1) loading blocks of inputs from HBM to SRAM, (2) computing attention with
respect to that block, and then (3) updating the output without writing the
large intermediate matrices $\vS$ and $\vP$ to HBM.
As the softmax couples entire rows or blocks of row, online
softmax~\citep{milakov2018online, rabe2021self} can split the attention
computation into blocks, and rescale the output of each block to finally get the
right result (with no approximation).
By significantly reducing the amount of memory reads/writes, \sysnameone yields
2-4$\times$ wall-clock speedup over optimized baseline attention implementations.

We describe the online softmax technique~\citep{milakov2018online} and how it is
used in attention~\citep{rabe2021self}.
For simplicity, consider just one row block of the attention matrix $\vS$, of the form
$\begin{bmatrix} \vS^{(1)} & \vS^{(2)} \end{bmatrix}$ for some matrices
$\vS^{(1)}, \vS^{(2)} \in \mathbb{R}^{B_r \times B_c}$, where $B_r$ and $B_c$ are the row and
column block sizes.
We want to compute softmax of this row block and multiply with the value,
of the form $\begin{bmatrix} \vV^{(1)} \\ \vV^{(2)} \end{bmatrix}$ for some
matrices $\vV^{(1)}, \vV^{(2)} \in \mathbb{R}^{B_c \times d}$.
Standard softmax would compute:
\begin{align*}
  m &= \max(\mathrm{rowmax}(\vS^{(1)}), \mathrm{rowmax}(\vS^{(2)})) \in \mathbb{R}^{B_r}  \\
  \ell &= \mathrm{rowsum}(e^{\vS^{(1)} - m}) + \mathrm{rowsum}(e^{\vS^{(2)} - m}) \in \mathbb{R}^{B_r}  \\
  \vP &= \begin{bmatrix} \vP^{(1)} & \vP^{(2)} \end{bmatrix} = \diag(\ell)^{-1}\begin{bmatrix} e^{\vS^{(1)} - m} & e^{\vS^{(2)} - m} \end{bmatrix} \in \mathbb{R}^{B_r \times 2B_c} \\
  \vO &= \begin{bmatrix} \vP^{(1)} & \vP^{(2)} \end{bmatrix} \begin{bmatrix} \vV^{(1)} \\ \vV^{(2)} \end{bmatrix} = \diag(\ell)^{-1} e^{\vS^{(1)} - m} \vV^{(1)} + e^{\vS^{(2)} - m} \vV^{(2)} \in \mathbb{R}^{B_r \times d}.
\end{align*}
Online softmax instead computes ``local'' softmax with respect to each block and
rescale to get the right output at the end:
\begin{align*}
  m^{(1)} &= \mathrm{rowmax}(\vS^{(1)})  \in \mathbb{R}^{B_r}\\
  \ell^{(1)} &= \mathrm{rowsum}(e^{\vS^{(1)} - m^{(1)}}) \in \mathbb{R}^{B_r} \\
  \tilde{\vP}^{(1)} &= \diag(\ell^{(1)})^{-1} e^{\vS^{(1)} - m^{(1)}} \in \mathbb{R}^{B_r \times B_c}\\
  \vO^{(1)} &= \tilde{\vP}^{(1)} \vV^{(1)} = \diag(\ell^{(1)})^{-1} e^{\vS^{(1)} - m^{(1)}} \vV^{(1)} \in \mathbb{R}^{B_r \times d}\\
  m^{(2)} &= \max(m^{(1)}, \mathrm{rowmax}(\vS^{(2)})) = m \\
  \ell^{(2)} &= e^{m^{(1)} - m^{(2)}} \ell^{(1)} + \mathrm{rowsum}(e^{\vS^{(2)} - m^{(2)}}) = \mathrm{rowsum}(e^{\vS^{(1)} - m}) + \mathrm{rowsum}(e^{\vS^{(2)} - m}) = \ell \\
  \tilde{\vP}^{(2)} &= \diag(\ell^{(2)})^{-1} e^{\vS^{(2)} - m^{(2)}} \\
  \vO^{(2)} &= \diag(\ell^{(1)} / \ell^{(2)})^{-1} \vO^{(1)} + \tilde{\vP}^{(2)} \vV^{(2)} = \diag(\ell^{(2)})^{-1} e^{s^{(1)} - m} \vV^{(1)} + \diag(\ell^{(2)})^{-1} e^{s^{(2)} - m} \vV^{(2)} = \vO.
\end{align*}

\begin{figure}[ht]
  \centering
  \includegraphics[width=0.9\textwidth]{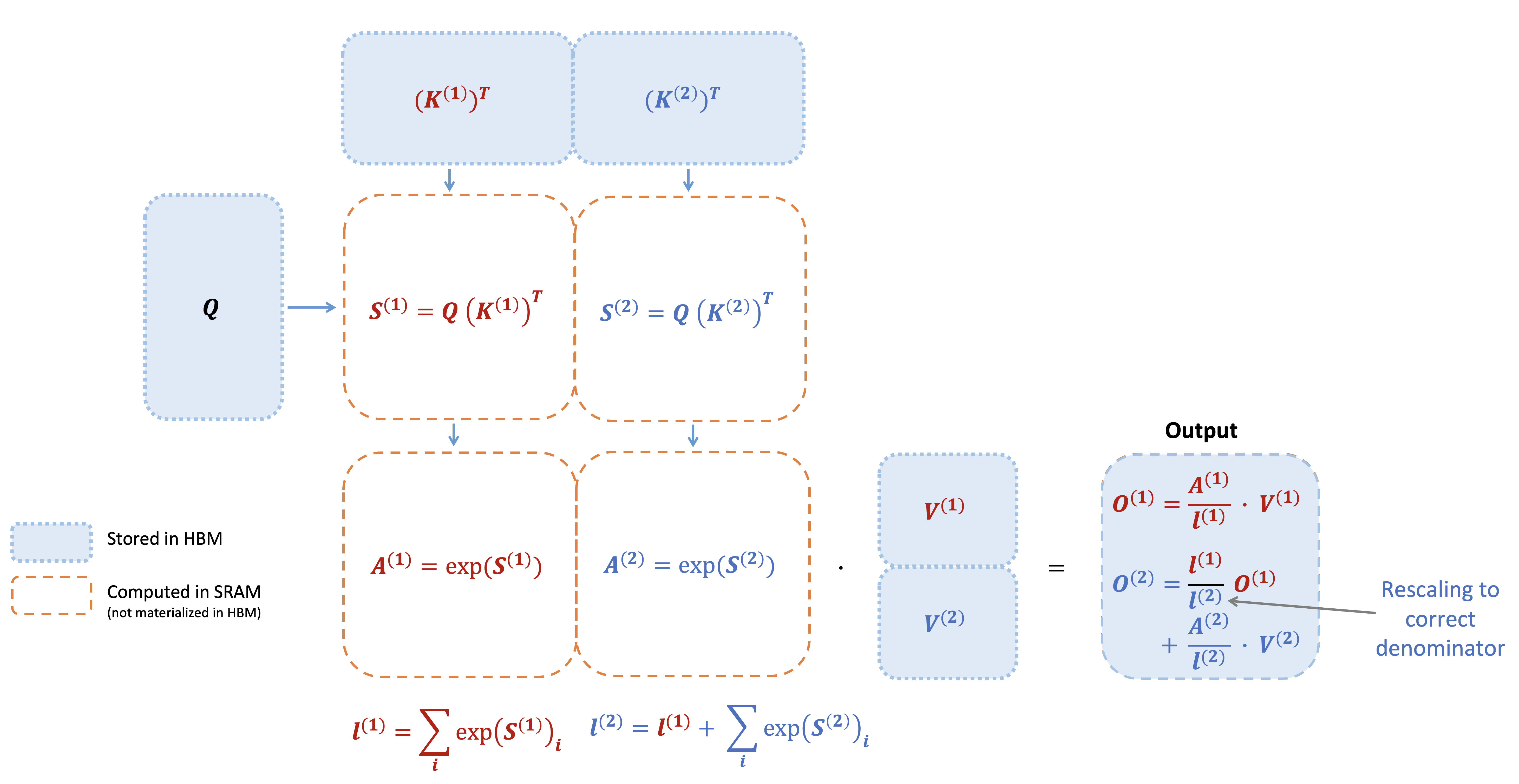}
  \caption{\label{fig:flash_attention_diagram}Diagram of how \sysnameone forward
    pass is performed, when the key $\vK$ is partitioned into two blocks and the
    value $\vV$ is also partitioned into two blocks.
    By computing attention with respect to each block and rescaling the output,
    we get the right answer at the end, while avoiding expensive memory
    reads/writes of the intermediate matrices $\vS$ and $\vP$.
    We simplify the diagram, omitting the step in softmax that subtracts each
    element by the row-wise max.}
\end{figure}

We show how \sysnameone uses online softmax to enable tiling
(\cref{fig:flash_attention_diagram}) to reduce memory reads/writes.

\subsubsection{Backward pass}
In the backward pass, by re-computing the values of the attention matrices $\vS$
and $\vP$ once blocks of inputs $\vQ, \vK, \vV$ are already loaded to SRAM,
\sysnameone avoids having to store large intermediate values.
By not having to save the large matrices $\vS$ and $\vP$ of size $N \times N$,
\sysnameone yields 10-20$\times$ memory saving depending on sequence length (memory
required in linear in sequence length $N$ instead of quadratic).
The backward pass also achieves 2-4$\times$ wall-clock speedup due to reduce memory
reads/writes.

The backward pass applies tiling to the equations in~\cref{subsec:standard_attn}.
Though the backward pass is simpler than the forward pass conceptually (there is
no softmax rescaling), the implementation is significantly more involved.
This is because there are more values to be kept in SRAM to perform 5 matrix
multiples in the backward pass, compared to just 2 matrix multiples in the
forward pass.


\section{\sysname: Algorithm, Parallelism, and Work Partitioning}
\label{sec:algo}

We describe the \sysname algorithm, which includes several tweaks to \sysnameone
to reduce the number of non-matmul FLOPs.
We then describe how to parallelize the computation on different thread blocks
to make full use the GPU resources.
Finally we describe we partition the work between different warps within one
thread block to reduce the amount of shared memory access.
These improvements lead to 2-3$\times$ speedup as validated in~\cref{sec:experiments}.

\subsection{Algorithm}
\label{subsec:algo}

We tweak the algorithm from \sysnameone to reduce the number of non-matmul
FLOPs.
This is because modern GPUs have specialized compute units (e.g., Tensor Cores
on Nvidia GPUs) that makes matmul much faster.
As an example, the A100 GPU has a max theoretical throughput of 312 TFLOPs/s of
FP16/BF16 matmul, but only 19.5 TFLOPs/s of non-matmul FP32.
Another way to think about this is that each non-matmul FLOP is 16$\times$ more
expensive than a matmul FLOP.
To maintain high throughput (e.g., more than 50\% of the maximum theoretical
TFLOPs/s), we want to spend as much time on matmul FLOPs as possible.

\subsubsection{Forward pass}

We revisit the online softmax trick as shown in~\cref{subsec:flashv1} and make
two minor tweaks to reduce non-matmul FLOPs:
\begin{enumerate}
  \item We do not have to rescale both terms of the output update by $\diag(\ell^{(2)})^{-1}$:
  \begin{equation*}
    \vO^{(2)} = \diag(\ell^{(1)} / \ell^{(2)})^{-1} \vO^{(1)} + \diag(\ell^{(2)})^{-1} e^{\vS^{(2)} - m^{(2)}} \vV^{(2)}.
  \end{equation*}
  We can instead maintain an ``un-scaled'' version of $\vO^{(2)}$ and keep
  around the statistics $\ell^{(2)}$:
  \begin{equation*}
    \tilde{\vO}^{(2)} = \diag(\ell^{(1)})^{-1} \vO^{(1)} + e^{\vS^{(2)} - m^{(2)}} \vV^{(2)}.
  \end{equation*}
  Only at the every end of the loop do we scale the final
  $\tilde{\vO}^{(\mathrm{last})}$ by $\diag(\ell^{(\mathrm{last})})^{-1}$ to get
  the right output.

  \item We do not have to save both the max $m^{(j)}$ and the sum of
  exponentials $\ell^{(j)}$ for the backward pass. We only need to store the
  logsumexp $L^{(j)} = m^{(j)} + \log(\ell^{(j)})$.
\end{enumerate}

In the simple case of 2 blocks in~\cref{subsec:flashv1}, the online softmax
trick now becomes:
\begin{align*}
  m^{(1)} &= \mathrm{rowmax}(\vS^{(1)})  \in \mathbb{R}^{B_r}\\
  \ell^{(1)} &= \mathrm{rowsum}(e^{\vS^{(1)} - m^{(1)}}) \in \mathbb{R}^{B_r} \\
  \tilde{\vO^{(1)}} &= e^{\vS^{(1)} - m^{(1)}} \vV^{(1)} \in \mathbb{R}^{B_r \times d}\\
  m^{(2)} &= \max(m^{(1)}, \mathrm{rowmax}(\vS^{(2)})) = m \\
  \ell^{(2)} &= e^{m^{(1)} - m^{(2)}} \ell^{(1)} + \mathrm{rowsum}(e^{\vS^{(2)} - m^{(2)}}) = \mathrm{rowsum}(e^{\vS^{(1)} - m}) + \mathrm{rowsum}(e^{\vS^{(2)} - m}) = \ell \\
  \tilde{\vP}^{(2)} &= \diag(\ell^{(2)})^{-1} e^{\vS^{(2)} - m^{(2)}} \\
  \tilde{\vO}^{(2)} &= \diag(e^{m^{(1)} - m^{(2)}})^{-1} \tilde{\vO}^{(1)} + e^{\vS^{(2)} - m^{(2)}} \vV^{(2)} = e^{s^{(1)} - m} \vV^{(1)} + e^{s^{(2)} - m} \vV^{(2)} \\
  \vO^{(2)} &= \diag(\ell^{(2)})^{-1} \tilde{\vO}^{(2)} = \vO.
\end{align*}

We describe the full \sysname forward pass in~\cref{alg:flash2_fwd}.

\begin{algorithm}[H]
  \caption{\small\label{alg:flash2_fwd}\sysname forward pass}
  \begin{algorithmic}[1]
    \REQUIRE Matrices $\vQ, \vK, \vV \in \mathbb{R}^{N \times d}$ in HBM, block sizes $B_c$, $B_r$.
    \STATE \label{alg:stream_attn_split_qkv} Divide $\vQ$ into $T_r = \left\lceil\frac{N}{B_r} \right\rceil$ blocks $\vQ_1, \dots, \vQ_{T_r}$ of size $B_r \times d$ each,
    and divide $\vK, \vV$ in to $T_c = \left\lceil \frac{N}{B_c} \right\rceil$ blocks $\vK_1, \dots, \vK_{T_c}$ and
    $\vV_1, \dots, \vV_{T_c}$, of size $B_c \times d$ each.
    \STATE Divide the output $\vO \in \mathbb{R}^{N \times d}$ into $T_r$ blocks $\vO_i, \dots, \vO_{T_r}$ of size
    $B_r \times d$ each, and divide the logsumexp $L$ into $T_r$ blocks $L_i, \dots, L_{T_r}$ of size
    $B_r$ each.
    \FOR{$1 \le i \le T_r$} \label{alg:stream_attn_outer_loop}
      \STATE \label{alg:stream_attn_load_q} Load $\vQ_i$ from HBM to on-chip SRAM.
      \STATE \label{alg:stream_attn_init} On chip, initialize $\vO_{i}^{(0)} = (0)_{B_r \times d} \in \mathbb{R}^{B_r \times d}, \ell_{i}^{(0)} = (0)_{B_r} \in \mathbb{R}^{B_r}, m_{i}^{(0)} = (-\infty)_{B_r} \in \mathbb{R}^{B_r}$.
      \FOR{$1 \le j \le T_c$}
        \STATE \label{alg:stream_attn_load_kv} Load $\vK_j, \vV_j$ from HBM to on-chip SRAM.
        \STATE \label{alg:stream_attn_qk} On chip, compute $\vS_{i}^{(j)} = \vQ_i \vK_j^T \in \mathbb{R}^{B_r \times B_c}$.
        \STATE \label{alg:stream_attn_statistics} On chip, compute
        $m_{i}^{(j)} = \mathrm{max}(m_{i}^{(j-1)}, \mathrm{rowmax}(\vS_{i}^{(j)})) \in \mathbb{R}^{B_r}$, $\tilde{\vP}_{i}^{(j)} = \exp(\vS_{i}^{(j)} - m_{i}^{(j)}) \in \mathbb{R}^{B_r \times B_c}$ (pointwise),
        $\ell_{i}^{(j)} = e^{m_{i}^{j-1} - m_{i}^{(j)}} \ell_{i}^{(j-1)} + \mathrm{row sum}(\tilde{\vP}_{i}^{(j)}) \in \mathbb{R}^{B_r}$.
        \STATE \label{alg:stream_attn_update} On chip, compute
        $\vO_{i}^{(j)} = \diag(e^{m_{i}^{(j-1)} - m_{i}^{(j)}})^{-1} \vO_{i}^{(j-1)} + \tilde{\vP}_{i}^{(j)} \vV_j$.
      \ENDFOR
      \STATE On chip, compute $\vO_{i} = \diag(\ell_{i}^{(T_c)})^{-1} \vO_{i}^{(T_c)}$.
      \STATE On chip, compute $L_{i} = m_{i}^{(T_c)} + \log(\ell_i^{(T_c)})$.
      \STATE Write $\vO_{i}$ to HBM as the $i$-th block of $\vO$.
      \STATE Write $L_{i}$ to HBM as the $i$-th block of $L$.
    \ENDFOR
    \STATE Return the output $\vO$ and the logsumexp $L$.
  \end{algorithmic}
\end{algorithm}

\paragraph{Causal masking.}

One common use case of attention is in auto-regressive language modeling, where
we need to apply a causal mask to the attention matrix $\vS$ (i.e., any entry
$\vS_{ij}$ with $j > i$ is set to $-\infty$).
\begin{enumerate}
  \item As \sysnameone and \sysname already operate by blocks, for any blocks
  where all the column indices are more than the row indices (approximately half
  of the blocks for large sequence length), we can skip the computation of that
  block.
  This leads to around 1.7-1.8$\times$ speedup compared to attention without the
  causal mask.
  \item We do not need to apply the causal mask for blocks whose row indices are
  guaranteed to be strictly less than the column indices. This means that for
  each row, we only need apply causal mask to 1 block (assuming square block).
\end{enumerate}

\paragraph{Correctness, runtime, and memory requirement.}
As with \sysnameone, \cref{alg:flash2_fwd} returns the correct output
$\vO = \softmax(\vQ\vK^\top)\vV$ (with no approximation), using $O(N^2d)$ FLOPs and
requires $O(N)$ additional memory beyond inputs and output (to store the
logsumexp $L$).
The proof is almost the same as the proof of
\citet[Theorem 1]{dao2022flashattention}, so we omit it here.

\subsubsection{Backward pass}

The backward pass of \sysname is almost the same as that of \sysnameone. We make
a minor tweak to only use the row-wise logsumexp $L$ instead of both the
row-wise max and row-wise sum of exponentials in the softmax.
We include the backward pass description in~\cref{alg:flash_bwd} for completeness.
\begin{algorithm}[h]
  \caption{\small\label{alg:flash_bwd}\sysname Backward Pass}
  \begin{algorithmic}[1]
    \REQUIRE Matrices $\vQ, \vK, \vV, \vO, \vdO \in \mathbb{R}^{N \times d}$ in HBM,
    vector $L \in \mathbb{R}^N$ in HBM, block sizes $B_c$, $B_r$.
    \STATE Divide $\vQ$ into $T_r = \left\lceil\frac{N}{B_r} \right\rceil$ blocks $\vQ_1, \dots, \vQ_{T_r}$ of size $B_r \times d$ each,
    and divide $\vK, \vV$ in to $T_c = \left\lceil \frac{N}{B_c} \right\rceil$ blocks $\vK_1, \dots, \vK_{T_c}$ and
    $\vV_1, \dots, \vV_{T_c}$, of size $B_c \times d$ each.
    \STATE Divide $\vO$ into $T_r$ blocks $\vO_i, \dots, \vO_{T_r}$ of size
    $B_r \times d$ each, divide $\vdO$ into $T_r$ blocks $\vdO_i, \dots, \vdO_{T_r}$
    of size $B_r \times d$ each, and divide $L$ into $T_r$ blocks $L_i, \dots, L_{T_r}$ of size
    $B_r$ each.
    \STATE Initialize $\vdQ = (0)_{N \times d}$ in HBM and divide it into $T_r$ blocks $\vdQ_1, \dots, \vdQ_{T_r}$ of size $B_r \times d$ each.
    Divide $\vdK, \vdV \in \mathbb{R}^{N \times d}$ in to $T_c$ blocks $\vdK_1, \dots, \vdK_{T_c}$ and
    $\vdV_1, \dots, \vdV_{T_c}$, of size $B_c \times d$ each.
    \STATE Compute $D = \mathrm{rowsum}(\vdO \circ \vO) \in \mathbb{R}^d$ (pointwise multiply), write
    $D$ to HBM and divide it into $T_r$ blocks $D_1, \dots, D_{T_r}$ of size
    $B_r$ each.
    \FOR{$1 \le j \le T_c$}
      \STATE Load $\vK_j, \vV_j$ from HBM to on-chip SRAM.
      \STATE Initialize $\vdK_j = (0)_{B_c \times d}, \vdV_j = (0)_{B_c \times d}$ on SRAM.
      \FOR{$1 \le i \le T_r$}
        \STATE Load $\vQ_i, \vO_i, \vdO_i, \vdQ_i, L_i, D_i$ from HBM to on-chip SRAM.
        \STATE On chip, compute $\vS_{i}^{(j)} = \vQ_i \vK_j^T \in \mathbb{R}^{B_r \times B_c}$.
        \STATE On chip, compute $\vP_{i}^{(j)} = \exp(\vS_{ij} - L_{i}) \in \mathbb{R}^{B_r \times B_c}$.
        \STATE On chip, compute
        $\vdV_j \leftarrow \vdV_j + (\vP_{i}^{(j)})^\top \vdO_i \in \mathbb{R}^{B_c \times d}$.
        \STATE On chip, compute
        $\vdP_{i}^{(j)} = \vdO_{i} \vV_j^\top \in \mathbb{R}^{B_r \times B_c}$.
        \STATE On chip, compute $\vdS_{i}^{(j)} = \vP_{i}^{(j)} \circ (\vdP_{i}^{(j)} - D_i) \in \mathbb{R}^{B_r \times B_c}$.
        \STATE Load $\vdQ_i$ from HBM to SRAM, then on chip, update
        $\vdQ_{i} \leftarrow \vdQ_i + \vdS_{i}^{(j)} \vK_j \in \mathbb{R}^{B_r \times d}$, and write
        back to HBM.
        \STATE On chip, compute $\vdK_{j} \leftarrow \vdK_j + {\vdS_{i}^{(j)}}^\top \vQ_i \in \mathbb{R}^{B_c \times d}$.
      \ENDFOR
      \STATE Write $\vdK_j, \vdV_j$ to HBM.
    \ENDFOR
    \STATE Return $\vdQ, \vdK, \vdV$.
  \end{algorithmic}
\end{algorithm}

\paragraph{Multi-query attention and grouped-query attention.}
Multi-query attention (MQA)~\citep{shazeer2019fast} and grouped-query attention
(GQA)~\citep{ainslie2023gqa} are variants of attention where multiple heads of
query attend to the same head of key and value, in order to reduce the size of
KV cache during inference.
Instead of having to duplicate the key and value heads for the computation, we
implicitly manipulate the indices into the head to perform the same computation.
In the backward pass, we need to sum the gradients $\vdK$ and $\vdV$ across
different heads that were implicitly duplicated.

\subsection{Parallelism}
\label{subsec:parallelism}

The first version of \sysnameone parallelizes over batch size and number of
heads.
We use 1 thread block to process one attention head, and there are overall
$\text{batch size} \cdot \text{number of heads}$ thread blocks.
Each thread block is scheduled to run on a streaming multiprocessor (SM), and
there are 108 of these SMs on an A100 GPU for example.
This scheduling is efficient when this number is large (say $\geq 80$), since we
can effectively use almost all of the compute resources on the GPU.

In the case of long sequences (which usually means small batch sizes or small
number of heads), to make better use of the multiprocessors on the GPU, we now
additionally parallelize over the sequence length dimension.
This results in significant speedup for this regime.

\paragraph{Forward pass.}
We see that the outer loop (over sequence length) is embarrassingly parallel,
and we schedule them on different thread blocks that do not need to communicate
with each other.
We also parallelize over the batch dimension and number of heads dimension, as
done in \sysnameone.
The increased parallelism over sequence length helps improve occupancy (fraction
of GPU resources being used) when the batch size and number of heads are small,
leading to speedup in this case.

These ideas of swapping the order of the loop (outer loop over row blocks and
inner loop over column blocks, instead of the other way round in the original
\sysnameone paper), as well as parallelizing over the sequence length dimension
were first suggested and implemented by Phil Tillet in the
Triton~\citep{tillet2019triton}
implementation.\footnote{\url{https://github.com/openai/triton/blob/main/python/tutorials/06-fused-attention.py}}

\paragraph{Backward pass.}
Notice that the only shared computation between different column blocks is in
update $\vdQ$ in \cref{alg:flash_bwd}, where we need to load $\vdQ_i$ from HBM
to SRAM, then on chip, update
$\vdQ_{i} \leftarrow \vdQ_i + \vdS_{i}^{(j)} \vK_j$, and write back to HBM.
We thus parallelize over the sequence length dimension as well, and schedule 1
thread block for each column block of the backward pass.
We use atomic adds to communicate between different thread blocks to update $\vdQ$.

\begin{figure}[ht]
  \centering
  \includegraphics[width=0.95\linewidth]{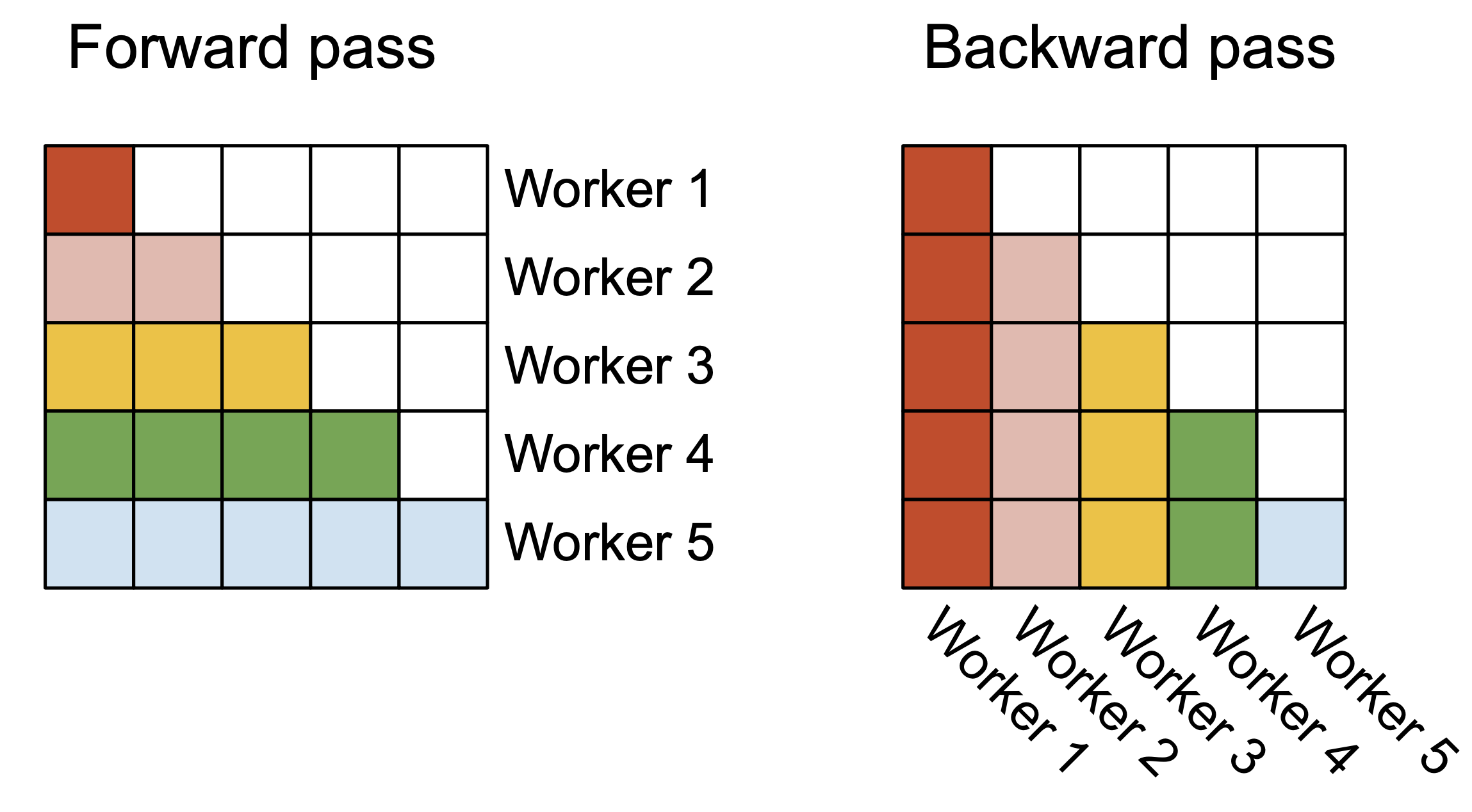}
  \caption{\label{fig:parallelism}In the forward pass (left), we parallelize the
    workers (thread blocks) where each worker takes care of a block of rows of
    the attention matrix.
    In the backward pass (right), each worker takes care of a block of columns
    of the attention matrix.}
\end{figure}

We describe the parallelization scheme in \cref{fig:parallelism}.

\subsection{Work Partitioning Between Warps}
\label{subsec:work_partitioning}

\begin{figure}[ht]
  \centering
  \begin{subfigure}{.53\textwidth}
    \centering
    \includegraphics[width=.95\linewidth]{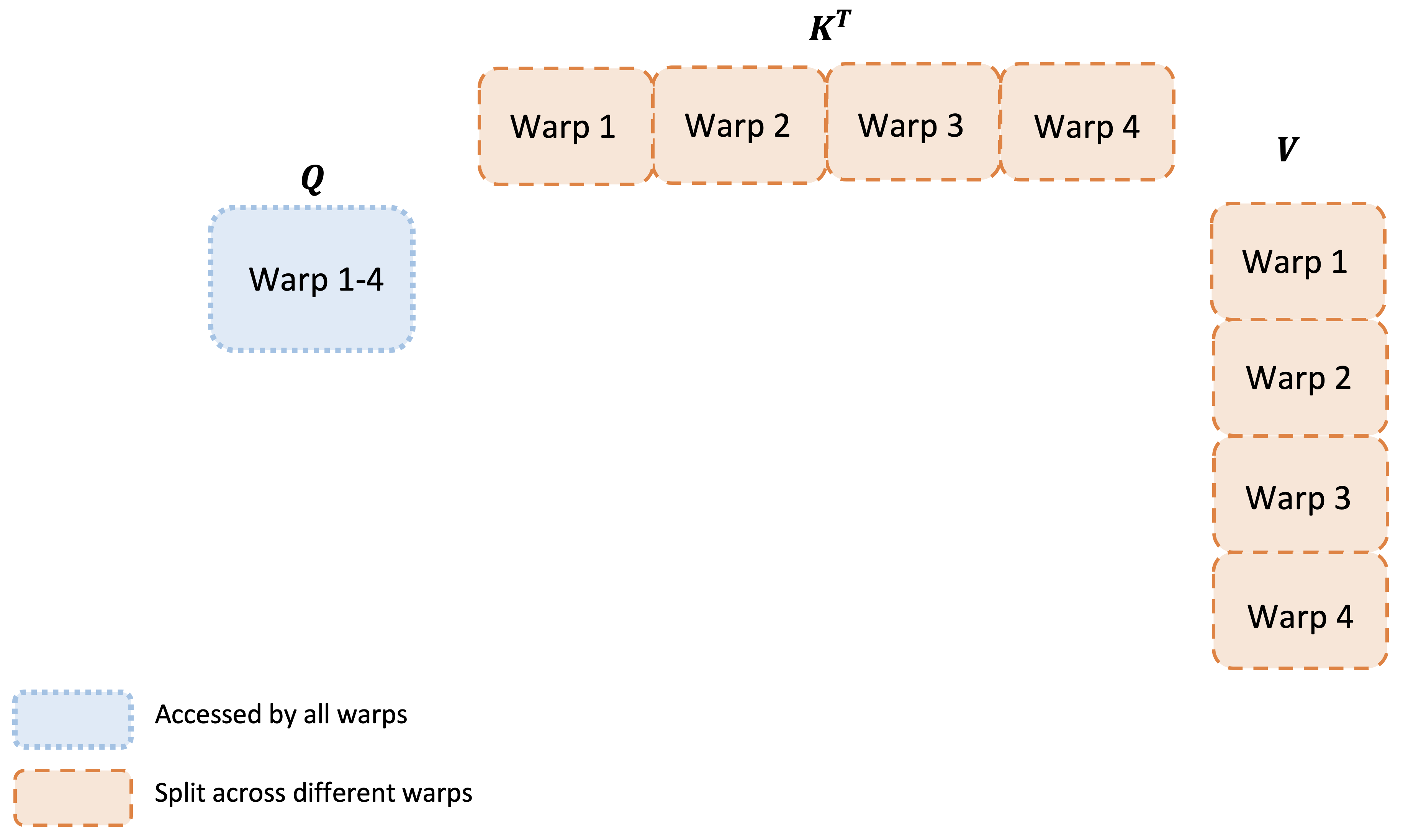}
    \caption{\sysnameone}
  \end{subfigure}%
  \begin{subfigure}{.47\textwidth}
    \centering
    \includegraphics[width=.95\linewidth]{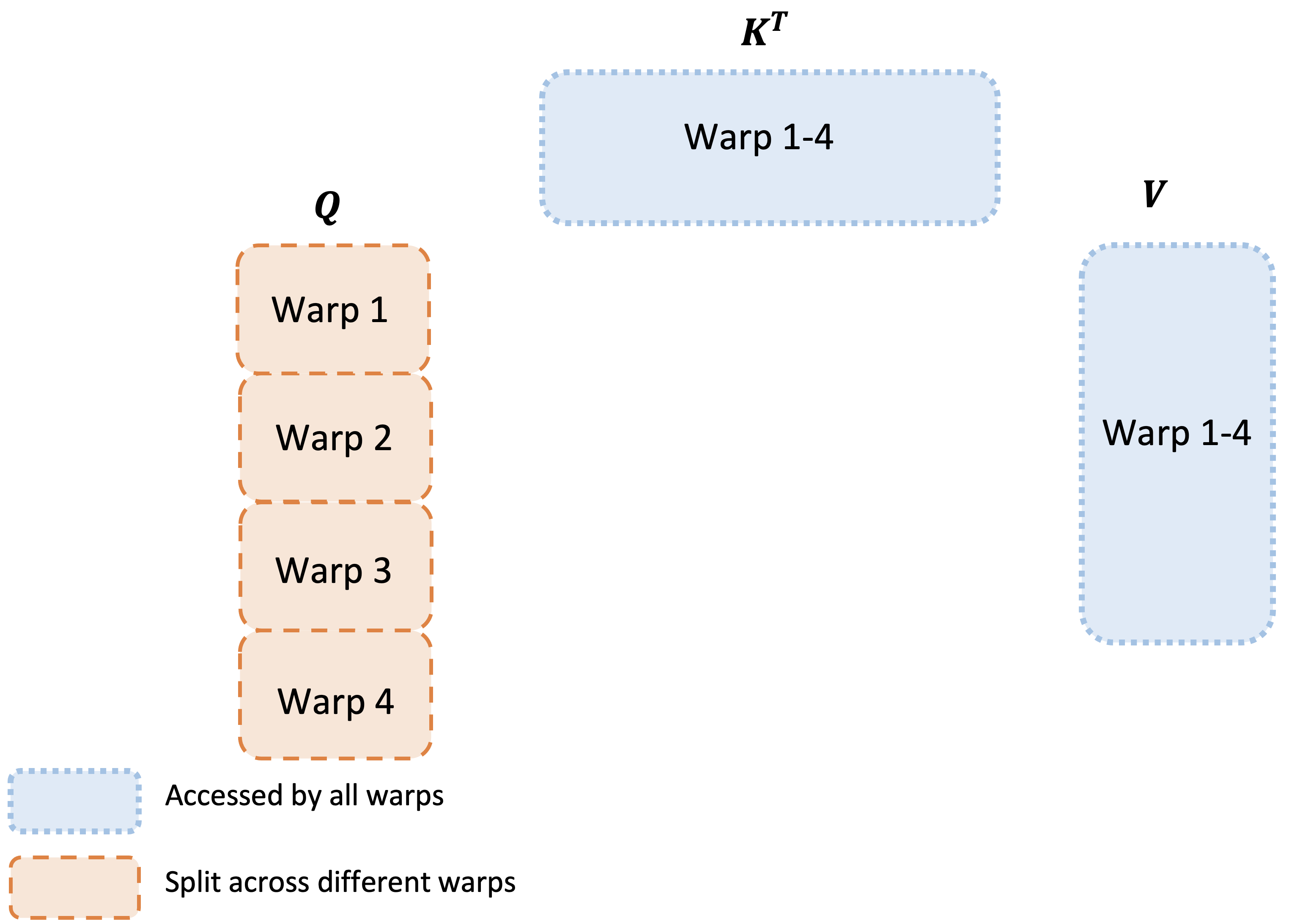}
    \caption{\sysname}
  \end{subfigure}
  \caption{Work partitioning between different warps in the forward pass}
  \label{fig:partitioning}
\end{figure}

As \cref{subsec:parallelism} describe how we schedule thread blocks, even within
each thread block, we also have to decide how to partition the work between
different warps.
We typically use 4 or 8 warps per thread block, and the partitioning is described
in \cref{fig:partitioning}.

\paragraph{Forward pass.}
For each block, \sysnameone splits $\vK$ and $\vV$ across 4 warps while keeping
$\vQ$ accessible by all warps.
Each warp multiplies to get a slice of $\vQ \vK^\top$, then they need to multiply
with a slice of $\vV$ and communicate to add up the result.
This is referred to as the ``split-K'' scheme.
However, this is inefficient since all warps need to write their intermediate
results out to shared memory, synchronize, then add up the intermediate results.
These shared memory reads/writes slow down the forward pass in \sysnameone.

In \sysname, we instead split $\vQ$ across 4 warps while keeping $\vK$ and $\vV$
accessible by all warps.
After each warp performs matrix multiply to get a slice of $\vQ \vK^\top$, they just need to
multiply with their shared slice of $\vV$ to get their corresponding
slice of the output.
There is no need for communication between warps.
The reduction in shared memory reads/writes yields speedup (\cref{sec:experiments}).

\paragraph{Backward pass.}
Similarly for the backward pass, we choose to partition the warps to avoid the
``split-K'' scheme.
However, it still requires some synchronization due to the more complicated
dependency between all the different inputs and gradients
$\vQ, \vK, \vV, \vO, \vdO, \vdQ, \vdK, \vdV$.
Nevertheless, avoiding ``split-K'' reduces shared memory reads/writes and again
yields speedup (\cref{sec:experiments}).

\paragraph{Tuning block sizes}
Increasing block sizes generally reduces shared memory loads/stores, but
increases the number of registers required and the total amount of shared
memory.
Past a certain block size, register spilling causes significant slowdown, or the
amount of shared memory required is larger than what the GPU has available, and
the kernel cannot run at all.
Typically we choose blocks of size $\{64, 128\} \times \{64, 128\}$, depending on the
head dimension $d$ and the device shared memory size.

We manually tune for each head dimensions since there are essentially only 4
choices for block sizes, but this could benefit from auto-tuning to avoid this
manual labor.
We leave this to future work.

\section{Empirical Validation}
\label{sec:experiments}

We evaluate the impact of using \sysname to train Transformer models.
\begin{itemize}[itemsep=0.1pt,topsep=0pt,leftmargin=*]
  \item \textbf{Benchmarking attention.}
  We measure the runtime of \sysname across different sequence lengths and
  compare it to a standard implementation in PyTorch, \sysnameone, and
  \sysnameone in Triton.
  We confirm that \sysname is 1.7-3.0$\times$ faster than \sysnameone, 1.3-2.5$\times$
  faster than \sysnameone in Triton, and 3-10$\times$ faster than a standard
  attention implementation.
  \sysname reaches up to 230 TFLOPs/s, 73\% of the theoretical maximum TFLOPs/s
  on A100 GPUs.
  \item \textbf{End-to-end training speed}
  When used end-to-end to train GPT-style models of size 1.3B and 2.7B on
  sequence lengths either 2k or 8k, \sysname yields up to 1.3$\times$ speedup compared to
  \sysnameone and 2.8$\times$ speedup compared to a baseline without \sysnameone.
  \sysname reaches up to 225 TFLOPs/s (72\% model FLOPs utilization) per A100 GPU.
\end{itemize}

\subsection{Benchmarking Attention}
\label{subsec:benchmark_attn}

\begin{figure}[ht]
  \centering
  \begin{subfigure}{.5\textwidth}
    \centering
    \includegraphics[width=.95\linewidth]{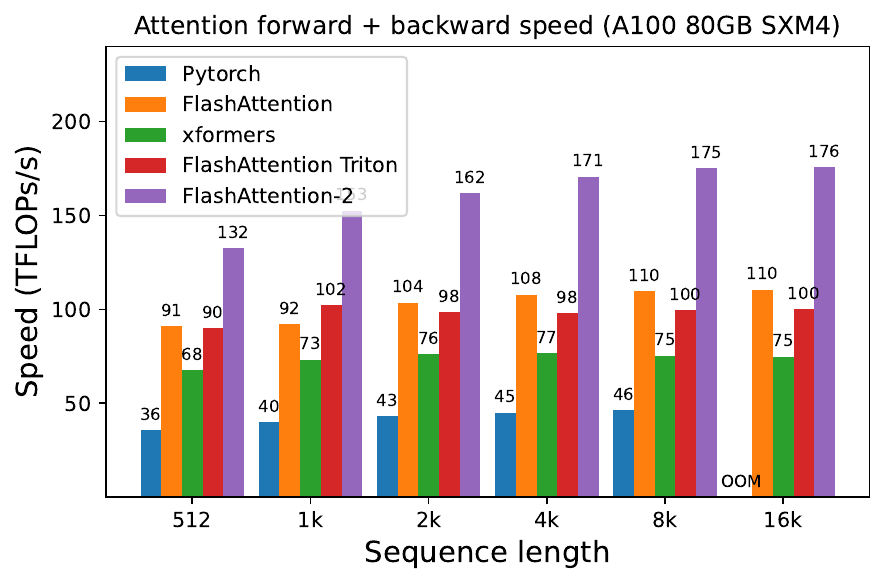}
    \caption{Without causal mask, head dimension 64}
  \end{subfigure}%
  \begin{subfigure}{.5\textwidth}
    \centering
    \includegraphics[width=.95\linewidth]{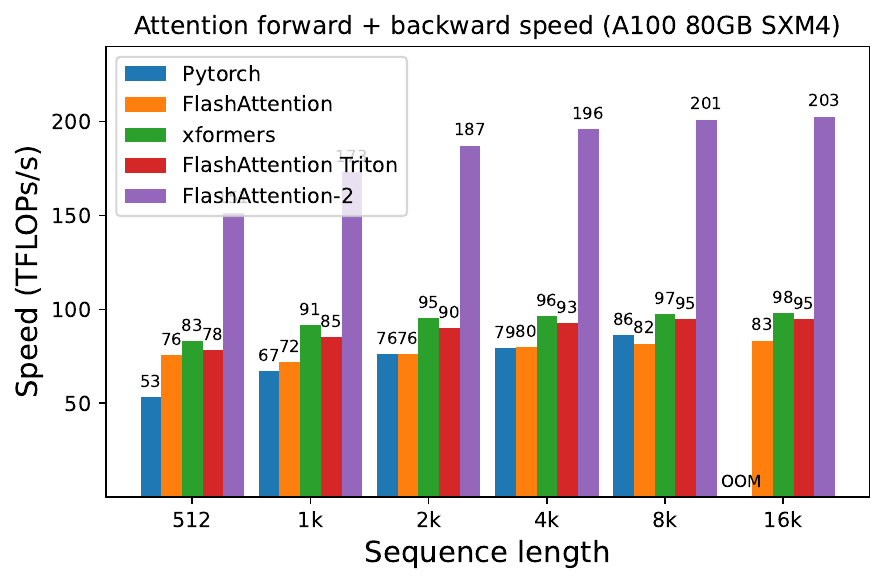}
    \caption{Without causal mask, head dimension 128}
  \end{subfigure}
  \begin{subfigure}{.5\textwidth}
    \centering
    \includegraphics[width=.95\linewidth]{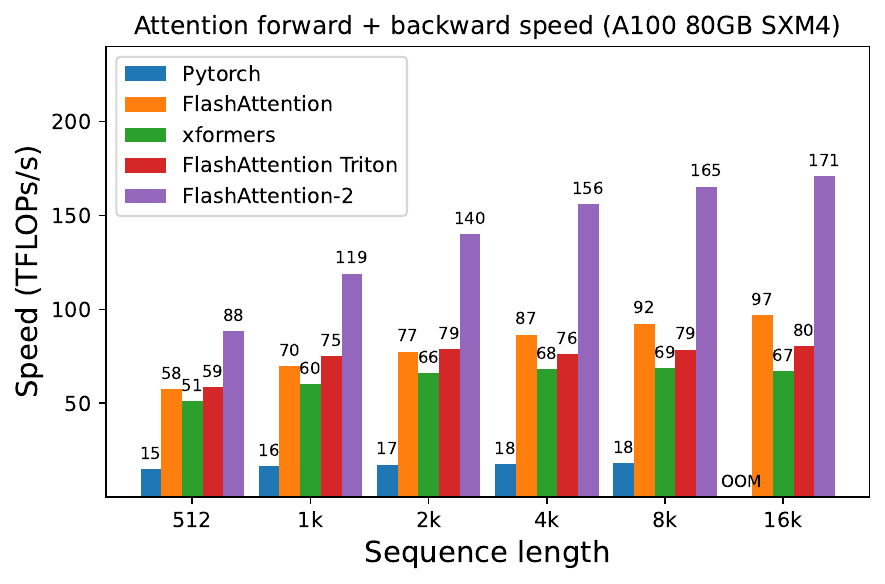}
    \caption{With causal mask, head dimension 64}
  \end{subfigure}%
  \begin{subfigure}{.5\textwidth}
    \centering
    \includegraphics[width=.95\linewidth]{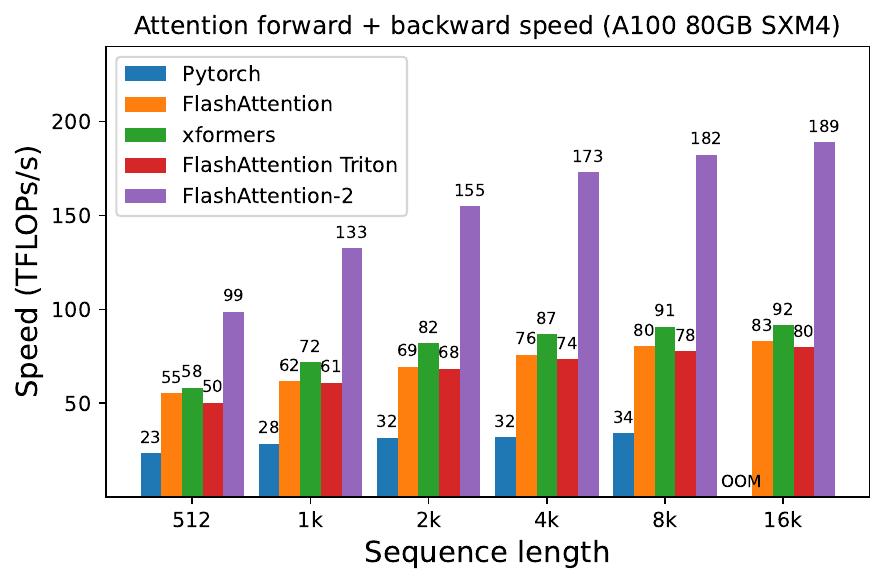}
    \caption{With causal mask, head dimension 128}
  \end{subfigure}
  \caption{Attention forward + backward speed on A100 GPU}
  \label{fig:benchmark_attn_fwd_bwd}
\end{figure}

\begin{figure}[ht]
  \centering
  \begin{subfigure}{.5\textwidth}
    \centering
    \includegraphics[width=.95\linewidth]{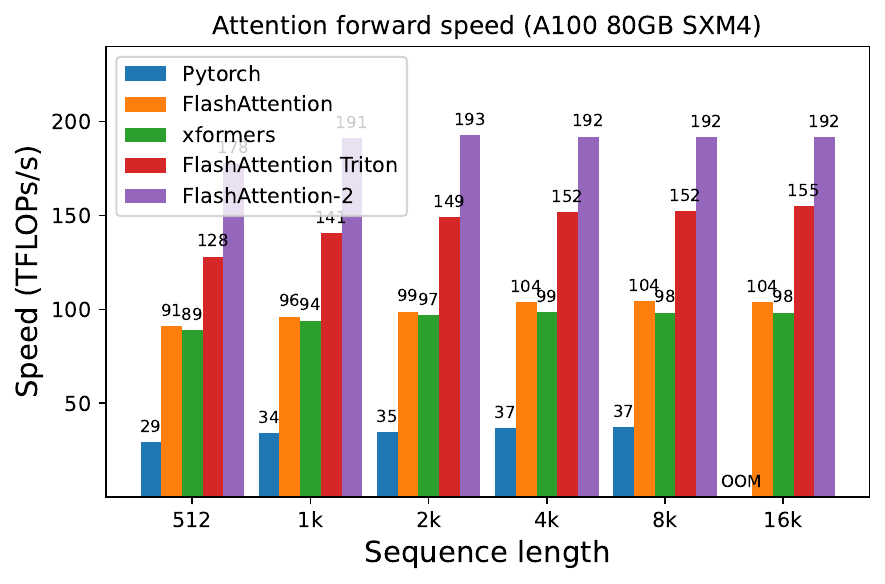}
    \caption{Without causal mask, head dimension 64}
  \end{subfigure}%
  \begin{subfigure}{.5\textwidth}
    \centering
    \includegraphics[width=.95\linewidth]{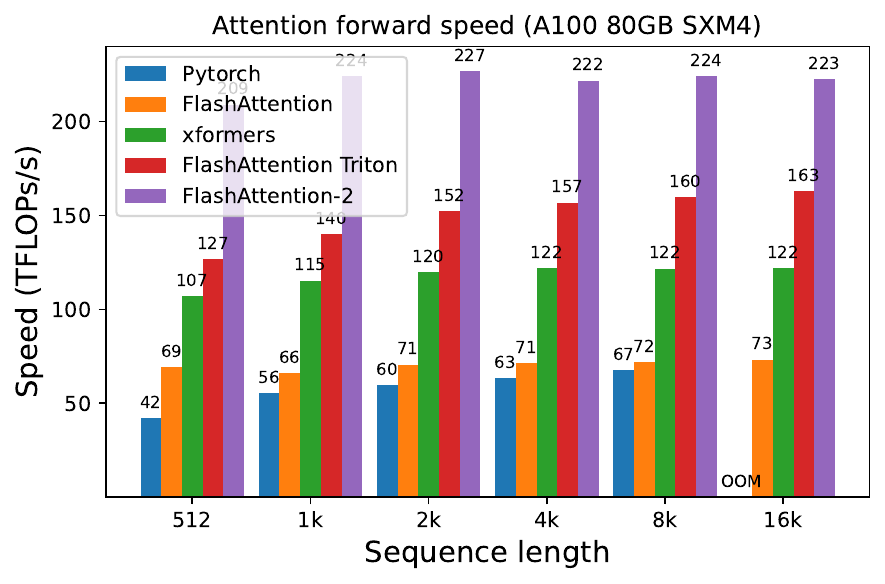}
    \caption{Without causal mask, head dimension 128}
  \end{subfigure}
  \begin{subfigure}{.5\textwidth}
    \centering
    \includegraphics[width=.95\linewidth]{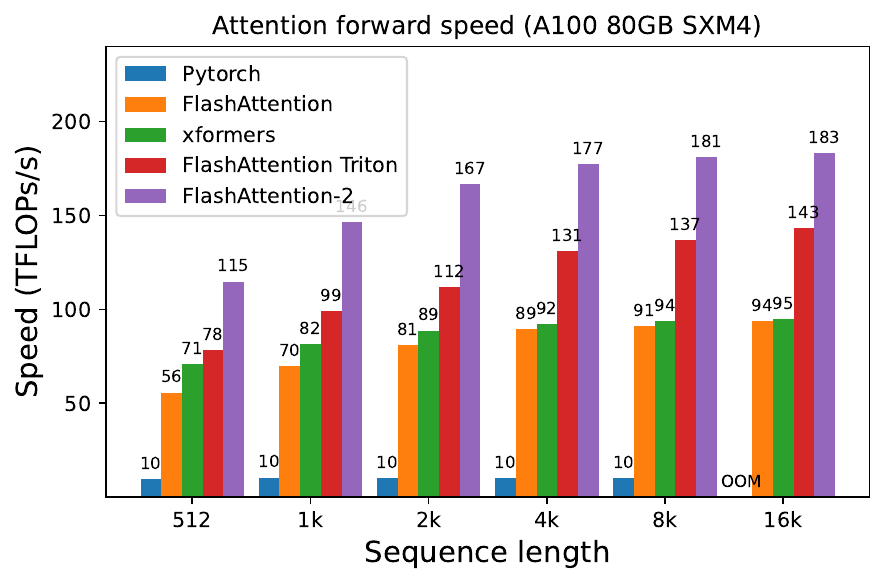}
    \caption{With causal mask, head dimension 64}
  \end{subfigure}%
  \begin{subfigure}{.5\textwidth}
    \centering
    \includegraphics[width=.95\linewidth]{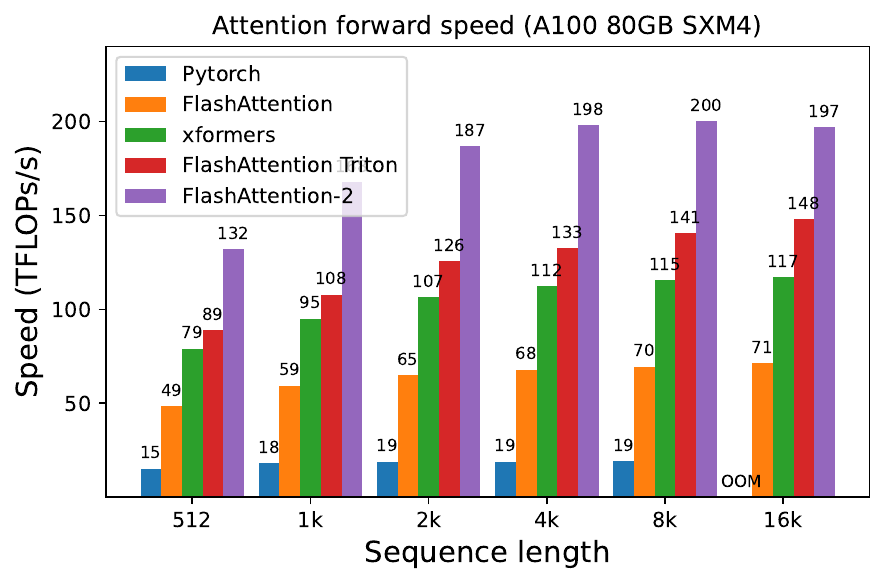}
    \caption{With causal mask, head dimension 128}
  \end{subfigure}
  \caption{Attention forward speed on A100 GPU}
  \label{fig:benchmark_attn_fwd}
\end{figure}

\begin{figure}[ht]
  \centering
  \begin{subfigure}{.5\textwidth}
    \centering
    \includegraphics[width=.95\linewidth]{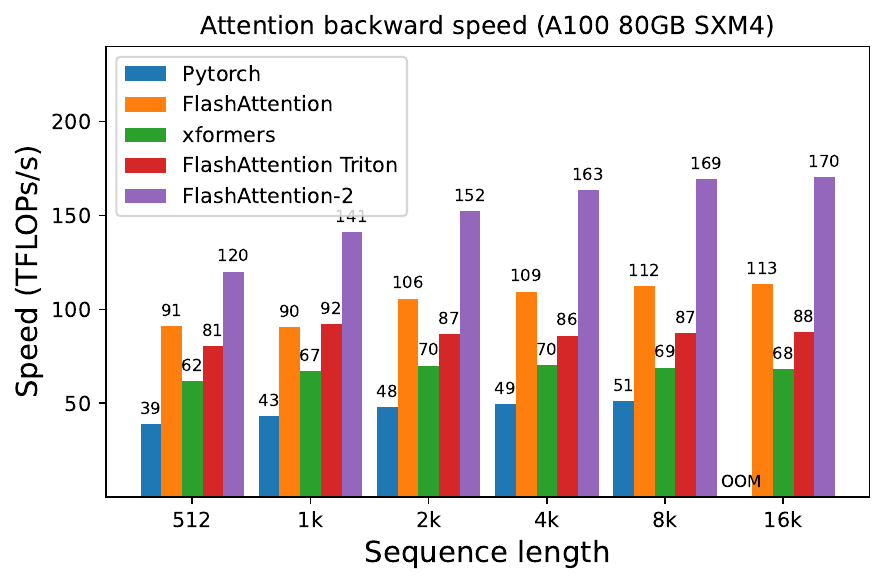}
    \caption{Without causal mask, head dimension 64}
  \end{subfigure}%
  \begin{subfigure}{.5\textwidth}
    \centering
    \includegraphics[width=.95\linewidth]{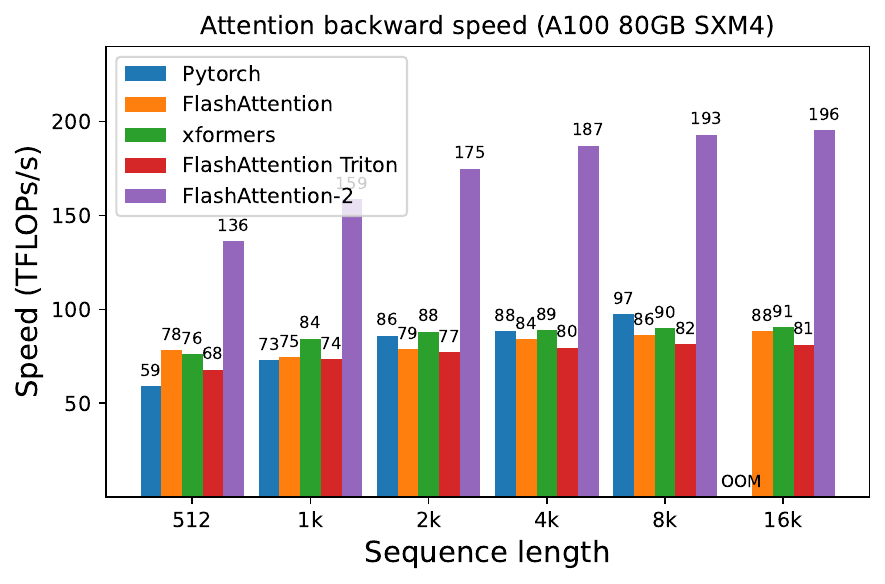}
    \caption{Without causal mask, head dimension 128}
  \end{subfigure}
  \begin{subfigure}{.5\textwidth}
    \centering
    \includegraphics[width=.95\linewidth]{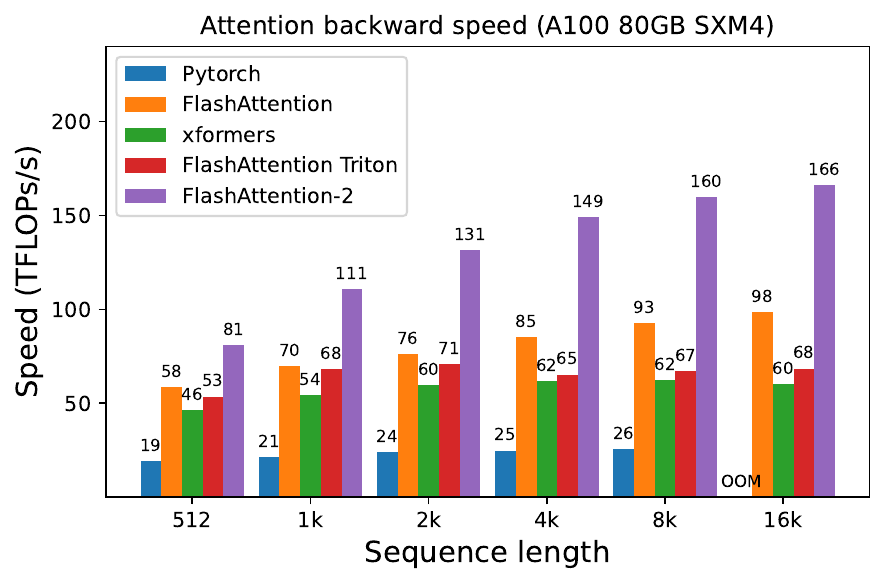}
    \caption{With causal mask, head dimension 64}
  \end{subfigure}%
  \begin{subfigure}{.5\textwidth}
    \centering
    \includegraphics[width=.95\linewidth]{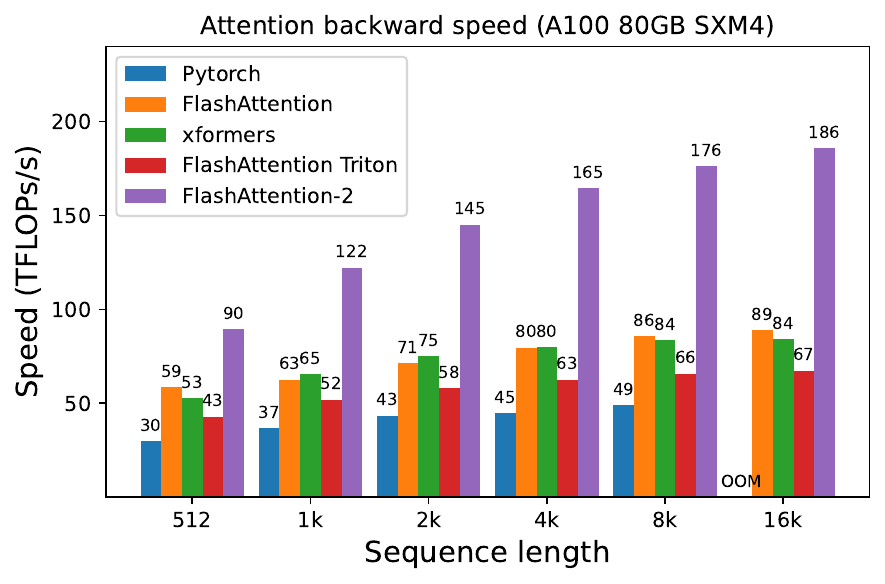}
    \caption{With causal mask, head dimension 128}
  \end{subfigure}
  \caption{Attention backward speed on A100 GPU}
  \label{fig:benchmark_attn_bwd}
\end{figure}

We measure the runtime of different attention methods on an A100 80GB SXM4 GPU
for different settings (without / with causal mask, head dimension 64 or 128).
We report the results
in~\cref{fig:benchmark_attn_fwd_bwd},~\cref{fig:benchmark_attn_fwd}
and~\cref{fig:benchmark_attn_bwd}, showing that \sysname is around 2$\times$ faster
than \sysnameone and \sysnameone in \texttt{xformers} (the ``cutlass''
implementation).
\sysname is around 1.3-1.5$\times$ faster than \sysnameone in Triton in the forward
pass and around 2$\times$ faster in the backward pass.
Compared to a standard attention implementation in PyTorch, \sysname can be up
to 10$\times$ faster.

Benchmark setting: we vary the sequence length from 512, 1k, ..., 16k, and set
batch size so that the total number of tokens is 16k.
We set hidden dimension to 2048, and head dimension to be either 64 or 128
(i.e., 32 heads or 16 heads).
To calculate the FLOPs of the forward pass, we use:
\begin{equation*}
  4 \cdot \text{seqlen}^2 \cdot \text{head dimension} \cdot \text{number of heads}.
\end{equation*}
With causal mask, we divide this number by 2 to account for the fact that
approximately only half of the entries are calculated.
To get the FLOPs of the backward pass, we multiply the forward pass FLOPs by 2.5
(since there are 2 matmuls in the forward pass and 5 matmuls in the backward
pass, due to recomputation).

\begin{figure}[ht]
  \centering
  \begin{subfigure}{.5\textwidth}
    \centering
    \includegraphics[width=.95\linewidth]{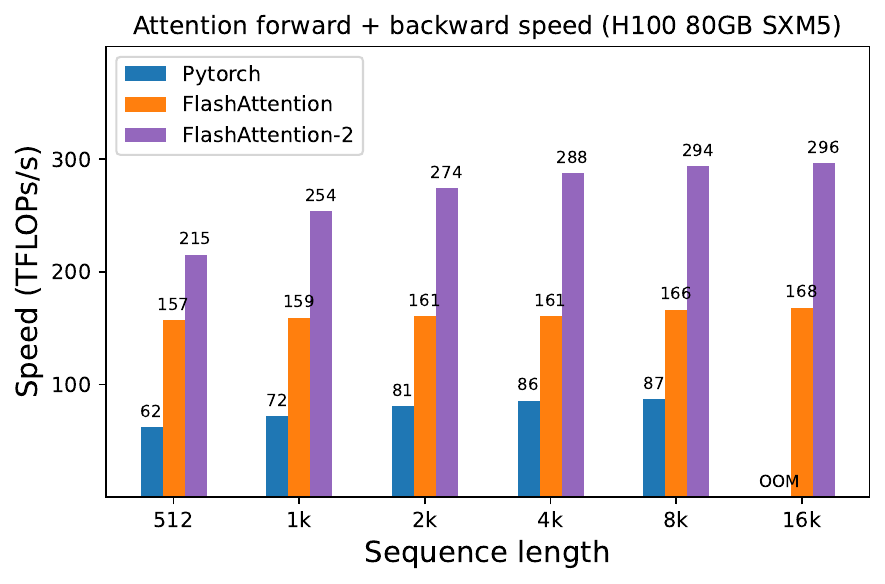}
    \caption{Without causal mask, head dimension 64}
  \end{subfigure}%
  \begin{subfigure}{.5\textwidth}
    \centering
    \includegraphics[width=.95\linewidth]{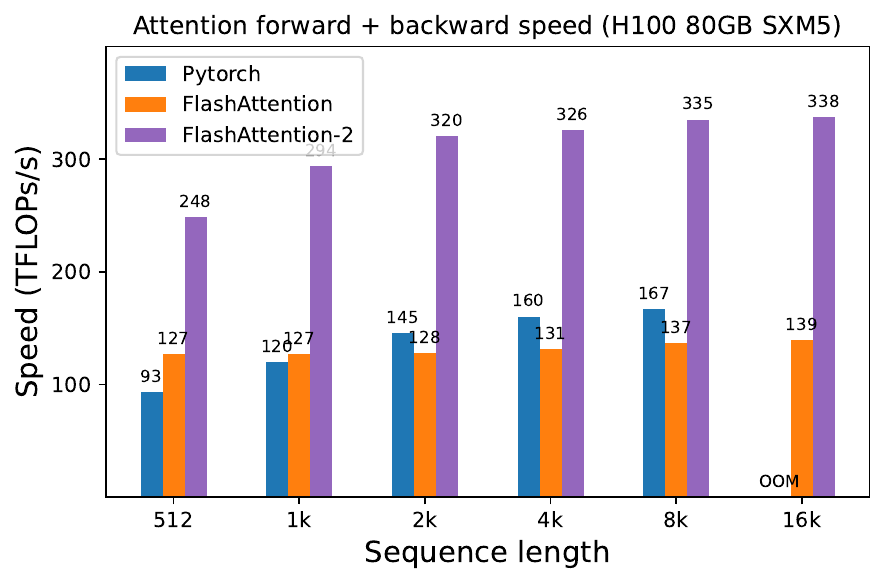}
    \caption{Without causal mask, head dimension 128}
  \end{subfigure}
  \begin{subfigure}{.5\textwidth}
    \centering
    \includegraphics[width=.95\linewidth]{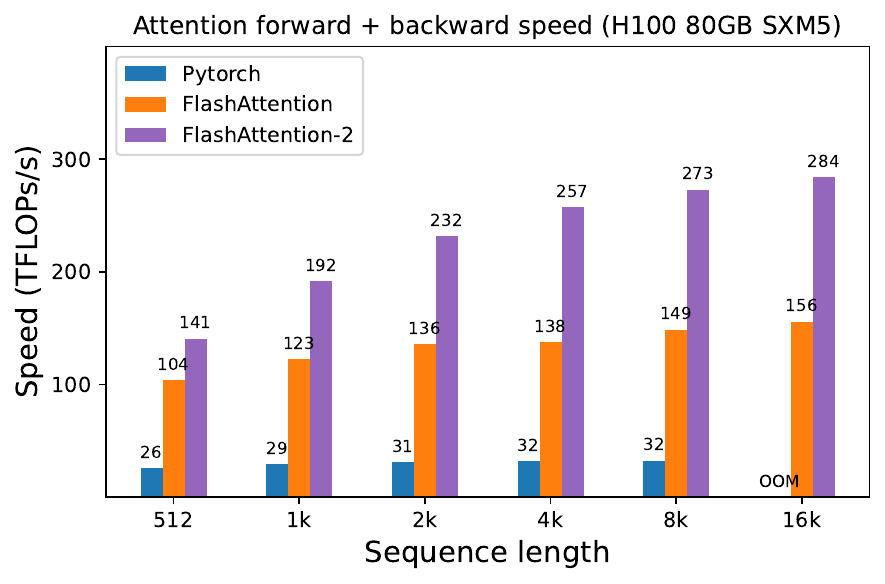}
    \caption{With causal mask, head dimension 64}
  \end{subfigure}%
  \begin{subfigure}{.5\textwidth}
    \centering
    \includegraphics[width=.95\linewidth]{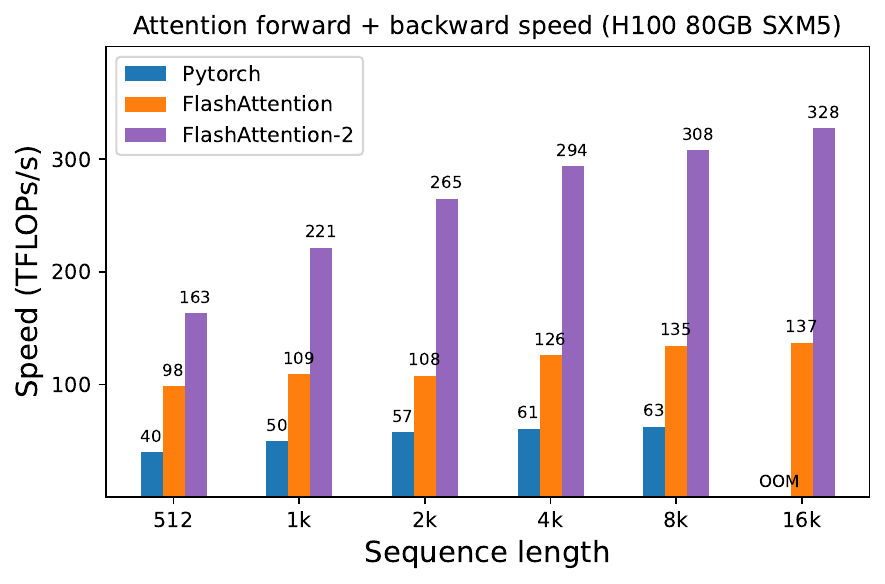}
    \caption{With causal mask, head dimension 128}
  \end{subfigure}
  \caption{Attention forward + backward speed on H100 GPU}
  \label{fig:benchmark_attn_fwd_bwd_h100}
\end{figure}

Just running the same implementation on H100 GPUs (using no special instructions
to make use of new features such as TMA and 4th-gen Tensor Cores), we obtain up
to 335 TFLOPs/s (\cref{fig:benchmark_attn_fwd_bwd_h100}).
We expect that by using new instructions, we can obtain another 1.5x-2x speedup
on H100 GPUs. We leave that to future work.

\subsection{End-to-end Performance}
\label{subsec:end_to_end}

We measure the training throughput of GPT-style models with either 1.3B or 2.7B
parameters, on 8$\times$A100 80GB SXM.
As shown in \cref{table:end_to_end}, \sysname yields 2.8$\times$ speedup compared to
a baseline without FlashAttention and 1.3$\times$ speedup compared to \sysname, reaching up to 225 TFLOPs/s
per A100 GPU.

Note that we calculate the FLOPs by the formula, following Megatron-LM~\citep{shoeybi2019megatron} (and many
other papers and libraries):
\begin{equation*}
  6 \cdot \text{seqlen} \cdot \text{number of params} + 12 \cdot \text{number of
    layers} \cdot \text{hidden dim} \cdot \text{seqlen}^2.
\end{equation*}
The first term accounts for the FLOPs due to weight--input multiplication, and
the second term accounts for the FLOPs due to attention.
However, one can argue that the second term should be halved, as with causal
mask we only need to compute approximately half the number of elements in
attention.
We choose to follow the formula from the literature (without dividing the
attention FLOPs by 2) for consistency.

\begin{table}[h!]
  \centering
  \caption{\label{table:end_to_end}Training speed (TFLOPs/s/GPU) of GPT-style
    models on 8$\times$A100 GPUs.
    \sysname reaches up to 225 TFLOPs/s (72\% model FLOPs utilization).
    We compare against a baseline running without \sysnameone.
  }
    {
      \begin{tabular}{@{}c|ccc@{}}
        Model & Without \sysnameone & \sysnameone & \sysname \\ \hline
        GPT3-1.3B 2k context & 142 TFLOPs/s & 189 TFLOPs/s & 196 TFLOPs/s \\
        GPT3-1.3B 8k context & 72 TFLOPS/s & 170 TFLOPs/s & 220 TFLOPs/s \\
        GPT3-2.7B 2k context & 149 TFLOPs/s & 189 TFLOPs/s & 205 TFLOPs/s \\
        GPT3-2.7B 8k context & 80 TFLOPs/s & 175 TFLOPs/s & 225 TFLOPs/s \\
      \end{tabular}
    }
  \end{table}


\section{Discussion and Future Directions}
\label{sec:discussion}

\sysname is 2$\times$ faster than \sysnameone, which means that we can train models
with 16k longer context for the same price as previously training a 8k context
model.
We are excited about how this can be used to understand long books and reports,
high resolution images, audio and video.
\sysname will also speed up training, finetuning, and inference of
existing models.

In the near future, we plan to collaborate with researchers and engineers to
make FlashAttention widely applicable in different kinds of devices (e.g., H100
GPUs, AMD GPUs), as well as new data types such as FP8.
As an immediate next step, we plan to optimize FlashAttention-2 for H100 GPUs to
use new hardware features (TMA, 4th-gen Tensor Cores, fp8).
Combining the low-level optimizations in FlashAttention-2 with high-level
algorithmic changes (e.g., local, dilated, block-sparse attention) could allow
us to train AI models with much longer context.
We are also excited to work with compiler researchers to make these optimization
techniques easily programmable.



\subsubsection*{Acknowledgments}

We thank Phil Tillet and Daniel Haziza, who have implemented versions of
\sysnameone in Triton~\citep{tillet2019triton} and the \texttt{xformers}
library~\citep{xFormers2022}.
\sysname was motivated by exchange of ideas between different ways that
attention could be implemented.
We are grateful to the Nvidia CUTLASS team (especially Vijay Thakkar, Cris Cecka, Haicheng
Wu, and Andrew Kerr) for their CUTLASS library, in particular the CUTLASS 3.x
release, which provides clean abstractions and powerful building blocks for the
implementation of \sysname.
We thank Driss Guessous for integrating \sysnameone to PyTorch.
\sysname has benefited from helpful discussions with Phil Wang, Markus Rabe,
James Bradbury, Young-Jun Ko, Julien Launay, Daniel Hesslow, Micha{\"e}l
Benesty, Horace He, Ashish Vaswani, and Erich Elsen.
Thanks for Stanford CRFM and Stanford NLP for the compute support.
We thank Dan Fu and Christopher R{\'e} for their collaboration, constructive
feedback, and constant encouragement on this line of work of designing
hardware-efficient algorithms.
We thank Albert Gu and Beidi Chen for their helpful suggestions on early drafts
of this technical report.

\bibliography{ref}
\bibliographystyle{plainnat}




\end{document}